\documentclass[]{spie}  

\usepackage{amsmath,amsfonts,amssymb}
\usepackage{graphicx}
\usepackage[colorlinks=true, allcolors=blue]{hyperref}
\usepackage{color,soul}
\usepackage[dvipsnames]{xcolor}
\usepackage{tabularray}
\usepackage{bm}

\title{Hyperspectral data augmentation with transformer-based diffusion models}

\author[a]{Mattia Ferrari}
\author[a]{Lorenzo Bruzzone}
\affil[a]{University of Trento, Via Calepina 14, Trento, Italy}

\authorinfo{Proc. SPIE 13196, Artifical Intelligence and Image Signal Processing Remote Sensing XXX, 131960I (2024), Preprint, Full version: \href{https://doi.org/10.1117/12.3032957}{10.1117/12.3032957}}

\begin{document} 
\maketitle

\begin{abstract}
The introduction of new generation hyperspectral satellite sensors, combined with advancements in deep learning methodologies, has significantly enhanced the ability to discriminate detailed land-cover classes at medium-large scales. However, a significant challenge in deep learning methods is the risk of overfitting when training networks with small labeled datasets. In this work, we propose a data augmentation technique that leverages a guided diffusion model. To effectively train the model with a limited number of labeled samples and to capture complex patterns in the data, we implement a lightweight transformer network. Additionally, we introduce a modified weighted loss function and an optimized cosine variance scheduler, which facilitate fast and effective training on small datasets. We evaluate the effectiveness of the proposed method on a forest classification task with 10 different forest types using hyperspectral images acquired by the PRISMA satellite. The results demonstrate that the proposed method outperforms other data augmentation techniques in both average and weighted average accuracy. The effectiveness of the method is further highlighted by the stable training behavior of the model, which addresses a common limitation in the practical application of deep generative models for data augmentation.
\end{abstract}


\section{INTRODUCTION}
\label{sec:intro} 
The way we analyze the Earth's surface has radically changed since the introduction of remote sensing (RS) sensors on aerial and satellite platforms. This new ability to observe and collect Earth's surface data has become essential over time for several reasons.
In forestry, RS is essential for mapping and monitoring forest cover, health, and composition. It is adopted for identifying tree species, estimating biomass, and detecting diseases, which are crucial variables for sustainable forest management and conservation efforts. Moreover, RS data are adopted to monitor deforestation and forest degradation, which are critical for understanding and mitigating the impacts of climate change. \cite{fassnacht2016review}

Forestry applications significantly benefit from the numerous spectral channels acquired by the last generation hyperspectral sensors. The availability of this type of data substantially enhance the ability to discriminate between land-cover objects and provide more precise measurements of many physical properties. The last generation hyperspectral sensors mounted on-board of satellites can acquire accurate spectral signatures, typically at wavelengths between 400-2500 nm, with a medium-high spatial resolution. An example is PRISMA \cite{guarini2017overview}, launched in 2019 by the Italian Space Agency (ASI), which is capable of acquiring 239 bands at 30m and 12 nm of spatial and spectral resolutions, respectively, and a panchromatic image at 5m of spatial resolution.

Along with the advancements in RS, land-cover classification accuracy has been significantly increased by the recent machine learning methodologies. In the last years, deep learning (DL) has made a substantial progress, enabling automatic and accurate classification of land cover using RS data. Traditional machine learning classification methods, such as Random Forest or Support Vector Machines, supported by hand-crafted feature extraction methods, have been used over the past decades to achieve good mapping results. However, these traditional methods are increasingly being replaced by DL methodologies, often based on convolutional neural networks (CNNs) and transformers networks. This change of paradigm is primarily due to the higher discrimination capability over different land cover classes of DL models, which is related to their higher ability to learn and discern complex patterns in the data. 

However, in DL the achievement of high classification accuracy is strongly dependent on the reliability and the quantity of the available labeled training samples. Since, the process of in-field surveys and ground truth data collection is often time-consuming, prone to errors, and costly, the quantity of data available to properly train DL networks is frequently limited and the quality affected by errors. This scarce data availability exacerbates the risk of overfitting reducing the generalization capability of the learned model that performs well on the training dataset but fails to generalize to new, unseen data, thereby limiting its practical utility. Moreover, the problem of overfitting is particularly pronounced in hyperspectral image classification due to the inherently high dimensionality of the data.

Several methods have been developed to address the overfitting issue in hyperspectral land cover classification. The most straightforward approach is to reduce the number of features using dimensionality reduction techniques, to decrease the complexity of the model and to reduce the amount of noise in the data. Among these methods, Principal Component Analysis (PCA) \cite{rodarmel2002principal} stands out for its simplicity and effectiveness with hyperspectral data. PCA reduces the dimensionality by transforming the original features into a new set of uncorrelated variables, capturing the most variance in the data. A more recent method involves the use of autoencoders to effectively reducing the number of features while learning the most informative representation of the data in the feature space. \cite{zabalza2016novel}

Another general approach to mitigate the risk of overfitting is to enhance the classification model by incorporating regularization techniques. These techniques help to penalize overly complex models, thus improving generalization and reducing overfitting. Common regularization methods include L1 regularization (Lasso) and L2 regularization (Ridge), which adds a penalty proportional to the absolute value and to the square of the magnitude of the coefficients, respectively. Another method is dropout, which reduces reliance on specific neurons by randomly deactivating a subset of neurons during each training iteration.

One of the best options to address the overfitting issue is data augmentation (DA). DA is a powerful yet straightforward technique that increases the quantity and diversity of training data, effectively overcoming the challenge of limited labeled samples in training DL models. This can be achieved by applying various transformations to the existing data, such as warping, scaling, and flipping, or by combining two or more samples to generate synthetic ones. These approaches help to introduce variability into the dataset, enhancing the ability of the DL model to generalize to unseen data. Moreover, DL generative models, such as Variational Autoencoders (VAEs) \cite{kingma2013auto} and Generative Adversarial Networks (GANs) \cite{goodfellow2014generative}, are employed to create entirely new samples that mimic the statistical properties of the original dataset. These synthetic data samples are crucial for mitigating overfitting, improving the robustness and generalization capabilities of the model.

However, the use of VAEs and GANs presents specific issues. VAE-generated data often suffers from low generation quality, as the models tend to produce samples that are not enough detailed due to the limitations in capturing complex distributions accurately. GANs, while capable of generating high-quality samples, face issues related to mode collapse, where the model learns to generate only a limited diversity of samples, and instability during training, which makes them difficult to train effectively.

Recently, diffusion generative probabilistic models (DMs) \cite{ho2020denoising} have emerged as a new category of generative DL models. DMs generate synthetic samples through a two-step process: initially adding noise to data samples and then learning to reverse this process to denoise the samples and recreate the original data distribution. DMs have become the state-of-the-art in computer vision tasks \cite{dhariwal2021diffusion}, since they are capable of producing high quality synthetic samples with a stable training process. However, they are still relatively underexploited in RS applications, being primarily explored for tasks like super-resolution and cloud detection \cite{liu2024diffusion}. However, the potential for these methods to generate high-quality, diverse synthetic hyperspectral data is promising yet remains relatively unexplored, as there is limited literature available on this topic.

In this work, we propose a novel DA method that generates synthetic hyperspectral samples using a guided DM architecture based on a transformer network. This method aims to overcome common challenges associated to the training of GAN models, while currently are the state-of-the-art methodologies in hyperspectral DA. The proposed method offers a more efficient and stable training process. Notably, we introduce modifications to the cosine scheduler and loss function, facilitating the effective training of the DM with a limited number of labeled samples. The introduced DM demonstrates the capability to generate high-quality spectral signatures, which can be incorporated into datasets to enhance classification accuracy and improve the training of DL networks.

The paper is structured as follows. The next section reviews various DA techniques for hyperspectral data, with particular emphasis on the DL generative models and the use of DMs with hyperspectral data. The Methodology section introduces DMs for sample generation and details our novel contributions to the proposed model. The Experimental Results section presents the outcomes of our experiments, comparing the proposed method with other techniques documented in the literature. Finally, the Conclusion section summarizes the key findings and provides concluding remarks.
\section{RELATED WORK}
\label{sec:relatedwork}
Various categories of Data Augmentation (DA) techniques for augmenting the number of labeled samples used in training DL models are proposed in the literature. The most basic methods rely on one-sample transformations. These transformations are extensively discussed in the literature, both in the spatial domain, using common techniques such as flipping and rotating single patches, and in the spectral domain, employing methods like scaling, adding noise, and warping. However, these approaches  often result in limited improvements in performance. \cite{li2018data}

A different category of approaches to sample augmentation involves mixing two or more samples. A commonly used method to address class imbalance in datasets, which can also be applied effectively to hyperspectral DA, is the Synthetic Minority Over-sampling TEchnique (SMOTE) \cite{chawla2002smote}. It creates synthetic samples by linearly interpolating between samples and their nearest neighbors belonging to the same class. 
Another example of mixing technique implements a data mixture model to augment the labeled training set and subsequently train a classifier on the augmented data  \cite{wang2019hyperspectral}. Then, by randomly sampling the coefficients in the data mixture model, several independent classifiers are generated and then combined using a voting strategy. 

Other approaches leverage the spatial context of samples for DA. For instance, a possible approach is to seek similar samples within the neighborhood to create sets of similar samples of varying window sizes \cite{wang2021data}. As the size of the sample set increases, the cluster center of the set converges toward the actual cluster center. These cluster centers, which reflect the distribution of the sample set, are then utilized as augmented samples for each current sample. 
An alternative technique selects some sample to be expanded into a pixel block to be used for training \cite{li2018data}. If any pixel block originates from the same class, its label remains unchanged; otherwise, a new label is assigned. During testing, the class label of each pixel is determined through a voting mechanism. 
Another method involves random occlusion of training samples from different rectangular spatial regions \cite{haut2019hyperspectral}. 
However, a significant limitation of this category of techniques in satellite hyperspectral RS is the typically low spatial resolution, which often renders neighboring pixels non-informative for classification purposes, thus restricting the applicability of these methods.

A different category of DA methods exploits deep generative models. The most used deep generative models in hyperspectral RS DA are GANs. Indeed, these models have demonstrated significant potential in generating realistic synthetic data that closely resembles the statistical properties of real hyperspectral samples.
The first proposed GAN for hyperspectral data is an Auxiliary Classifier-GAN (AC-GAN) composed by both generator and discriminator based on a convolutional neural network (CNN) with some changes to the objective function \cite{zhu2018generative}. 
An alternative strategy to AC-GAN is proposed in a framework which incorporates a transformer and an external semisupervised classifier \cite{he2022hypervitgan}. The model uses a generator and discriminator with skip connections to generate hyperspectral data.
Another example of GAN proposed for augmenting data in hyperspectral domain uses a Wasserstein GAN with a gradient penalty and introduces a classifier to condition the generator on classes  \cite{audebert2018generative}. The classification network adds a conditional penalty to ensure that the generated samples are classified according to the given class labels.
In a different study, a CramérGAN is adopted to enhance training stability and to improve the generated samples diversity \cite{hennessy2021generative}. 
Other methods exploit GANs for hyperspectral DA in a more sophisticated way. 

Another literature approach generates additional samples by preserving some spectral bands of the original samples through a band selection technique and generating the other spectral bands using a mixture strategy between real and GAN synthetically generated samples bands. \cite{zhang2023features} This ensures that the augmented data maintain the essential features of the original ones.
An alternative framework proposes a Triple Generative Adversarial Network (TripleGAN) for generating samples \cite{wang2019caps}. Notably, this model also incorporates Capsule Networks (CapsNets) for the classification. 

However, despite the several designs of GANs proposed in the literature partially overcoming the typical well-known issue of these networks, they anyway often face limitations such as mode collapse, where the model generates a limited variety of samples, failing to capture the full diversity of the training data. They also tend to require careful tuning of hyperparameters and can be unstable during training, leading to difficulties in achieving convergence. 

\subsection{Diffusion Models}

DMs, when compared to GANs, offer a notable advantage in terms of ease and stability of training. In the field of computer vision, numerous studies on DM have been published \cite{croitoru2023diffusion}, revealing results that surpass the performance of state-of-the-art GAN architectures. \cite{rombach2022high,dhariwal2021diffusion} In RS, and in particular with hyperspectral data, DMs are commonly used for super resolution, cloud removal and denoising \cite{liu2024diffusion}. In the domain of land-cover classification of hyperspectral data, the DMs are mainly adopted to extract features from the data. 
A first example of this approach consists of a spectral-spatial diffusion module and an attention-based classification module \cite{chen2023spectraldiff}. The primary focus of this architecture is to effectively exploit the distribution of hyperspectral data within high-dimensional and highly redundant data by iteratively denoising and explicitly constructing the data generation process.
Another method \cite{sigger2024unveiling} exploits forward and reverse diffusion processes to learn high-level and low-level features, extracting intermediate hierarchical features from a denoising U-Net at various timestamps, and then employing a transformer-based classifier. 
A different approach proposes a diffusion-based feature learning framework, which leverages multi-timestep, multi-stage diffusion features  \cite{zhou2024exploring}. The framework includes a class and timestep-oriented multi-stage feature purification (CTMSFP) module to reduce feature redundancy. It also exploits a selective timestep feature fusion module to adaptively integrate texture and semantic features, enhancing the generality and adaptability of the model across diverse hyperspectral data patterns. 

However, considering the literature in augmenting techniques developed for RS data using DMs, the majority of methods can be categorized as text-to-image or image-to-image generation \cite{liu2024diffusion}. In text-to-image generation, DMs have been adopted using trainable text-image pairs from existing RS images, incorporating numerical and feature-based prompts to enhance image synthesis. In image-to-image generation, DMs use guiding images like maps, or multi-modal RS images to generate new images. Nevertheless, if we consider purely DA techniques, very few techniques using DMs are presented for RS data. One of these methods is made-up of a four-stage approach aimed at improving the diversity of augmented data by integrating DMs \cite{sousa2024data}. The approach utilizes meta-prompts for instruction generation, leverages general-purpose vision-language models to produce detailed captions, fine-tunes an Earth observation DM, and iteratively augments the data. Another method \cite{wu2023high} exploits a traditional DM based on U-Net to generate synthetic samples in active deception jamming recognition. However, in the literature, no DA approaches based on DMs are specifically developed to generate synthetic hyperspectral samples.
\section{METHODOLOGY}
\label{sec:sections}
DMs represent a class of generative models that are designed to generate data by iteratively transforming an initial simple distribution, such as a Gaussian noise, into the complex distribution characteristic of the data to generate. This transformation is obtained by evolving the data distribution over time through a series of small incremental denoising steps. The framework for DMs can be mathematically articulated in terms of both a forward and a reverse process, which together define the dynamics of the model.

During the forward diffusion process, the model receives as input a data sample sampled from a real data distribution, denoted as $\mathbf{x}_0 \sim q(\mathbf{x})$, and progressively adds Gaussian noise $\bm{\epsilon}$ across $T$ discrete timesteps. At each timestep $t$, the data becomes increasingly noisy, losing its distinctive data distribution and becoming an isotropic Gaussian distribution at $T \to \infty$. The noise injected at each timestep $t$ is determined by a variance scheduler $\{\beta_t \in (0,1)\}^T_{t=1}$. 
The forward step can be formally defined as follows:
\begin{equation}
q(\mathbf{x}_t|\mathbf{x}_{t-1})=\mathcal{N}(\mathbf{x}_t;\sqrt{1-\beta_t}\mathbf{x}_{t-1},\beta_t\mathbf{I})
\end{equation}
The noised sample at timestep $t$ is expressed as:
\begin{equation}
\mathbf{x}_t = \sqrt{\bar{\alpha}_t} \mathbf{x}_0 + \sqrt{1 - \bar{\alpha}_t} \bm{\epsilon}, \quad \bm{\epsilon} \sim \mathcal{N}(0, I)
\label{eq:noising}
\end{equation}
where $\bar{\alpha}_t=\prod_{i=0}^{t}\alpha_i$, $\alpha_t = 1-\beta_t$. The role of the $\bar{\alpha}_t$ is to control the extent of noise added at each step, thereby enabling the data to transition smoothly from the clean sample to pure noise. 
A common and effective variance scheduler is the cosine scheduler \cite{nichol2021improved} that is defined as follows:
\begin{equation}
\beta_{t}=\text{clip}(1-\frac{\bar{\alpha_t}}{\bar{\alpha_{t-1}}},0.999)
\;\;\;\; 
\bar{\alpha_t}=\frac{f(t)}{f(0)}
\;\;\;\; 
\text{where}\;f(t)=\text{cos}(\frac{\frac{t}{T}+s}{1+s}\cdot\frac{\pi}{2})^\delta
\label{eq:var_sched}
\end{equation}
where $\delta = 2$ and $s$ is a small value to prevent $\beta_t$ from being too small when close to $t=0$.
This particular scheduler is introduced to replace the original proposed linear scheduler \cite{ho2020denoising} showing improvement in quality of the generated samples. 

In this work, we propose a modification to $\delta=1.2$, which empirically shows better network convergence during training with small values of $T$. This modification results in a smoother behavior of $\bar{\alpha}_t$ when $t$ is small, allowing for a smaller variance in the initial diffusion steps. As $t$ increases, the variance grows, nearly reaching an isotropic Gaussian distribution at $T$ timestep.

The core of DMs lies in the reverse process. It aims to denoise the data, effectively transforming the noised samples back to the original data distribution. However, the inversion of the forward process $q(\mathbf{x}_{t-1}|\mathbf{x}_t)$ is mathematically intractable, and so the reverse process is approximated with a parameterized model $p_\theta(\mathbf{x}_{t-1}|\mathbf{x}_t)$, estimating only the mean $\bm{\mu}_\theta$ and the variance $\bm{\Sigma}_\theta$ since for small $\beta_t$ the reverse process is also Gaussian $p_\theta(\mathbf{x}_{t-1}|\mathbf{x}_t)=\mathcal{N}(\mathbf{x}_{t-1};\bm{\mu}_\theta(\mathbf{x}_t,t),\bm{\Sigma}_\theta(\mathbf{x}_t,t))$. Additionally, Since, we want to generate samples belonging to specific labels, we add class conditioning information at each reverse diffusion step $p_\theta(\mathbf{x}_{t-1}|\mathbf{x}_t,y)$ to guide the reverse process.

\begin{figure}
\centering
\includegraphics[width=\textwidth]{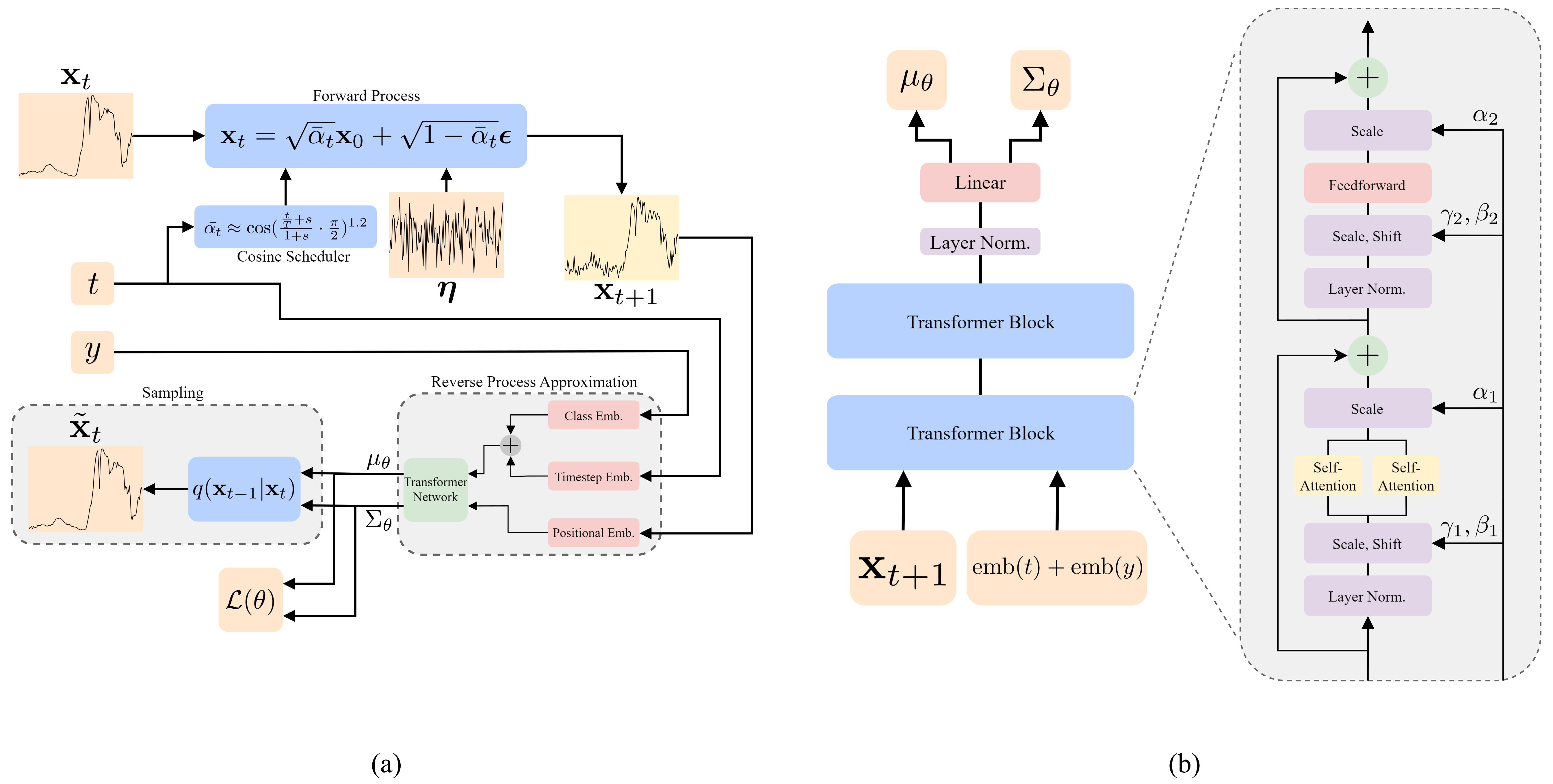}
\caption{Illustration of a diffusion step of the proposed model (a) and of the transformer model used to approximate the reverse process (b)}
\label{fig:net}
\end{figure}

Traditionally, the reverse process has been parameterized using U-Net architectures \cite{ho2020denoising}, which are well-suited for capturing local spatial correlations in data. However, in this work, inspired by Scalable DMs with transformers \cite{peebles2023scalable}, we adopt a transformer network. This is motivated by its powerful capability to model long-range dependencies and capture complex patterns in the data, particularly significant in hyperspectral data. The attention mechanisms of transformers allow them to weight different parts of the input data differently, making them particularly effective in handling the different data distributions of classes, which is crucial in the reverse diffusion process where fine details need to be recovered from noisy inputs. An overview of the entire proposed DM framework is shown in Figure \ref{fig:net}, while the design of the transformer model is depicted in Figure \ref{fig:net}.

The design of the model is guided by the requirements to have a small size model to allow a fast and effective approximation of the reverse process, but that should handle the complexity of the spectral signature pattern adequately. So, we choose to employ a transformer network with this dimensional characteristic: depth equal to two, two heads for each block and a hidden size of two. Considering the necessity of class guiding the reverse process approximation, the embedding of the timestep has been summed to the embedding of the class of the samples given as input, while the transformer block is similar to the one proposed by Peebles and Xie\cite{peebles2023scalable}. The key part of this block is the AdaLN-Zero block that combines adaptive layer normalization with a zero-initialization strategy. The first dynamically generates normalization parameters based on contextual embeddings, while the second is used for dimension-wise scaling parameters before residual connections to ensure each block starts as an identity function, thereby stabilizing and accelerating training.

The loss function used is the traditional one \cite{ho2020denoising}, with an additional loss weight $w$ inspired by Signal-to-Noise Ratio weighting with $\text{SNR}(t)=\frac{\bar{\alpha}_t}{1-\bar{\alpha}_t}$. This adjustment is necessary because we predict $\epsilon_\theta(t)$; in this context, using this weight results in consistent weighting when predicting $\mathbf{x}_{t-1}$. To give more importance to the initial diffusion steps, we propose modifying the SNR loss weighting as follows:
\begin{equation}
\mathcal{L(\theta)}=\underbrace{||\epsilon_\theta(t)-\epsilon_t||^2_{2}}_\text{MSE}w+\underbrace{\sum_{t}\mathcal{D}_{KL}(q(\mathbf{x}_{t-1}|\mathbf{x}_t,\mathbf{x}_0)||p_\theta(\mathbf{x}_{t-1}|\mathbf{x}_t,y))}_\text{VLB} 
\; \; \; \text{where} \; w=\text{norm}(\frac{\text{SNR}(t)}{\text{SNR}(t)\gamma+1})
\label{eq:loss eq}
\end{equation}
More in detail, the loss function in Eq. \ref{eq:loss eq} is composed by two parts: the first part is represented by the mean square error (MSE) between the noise predicted by the model $\epsilon_\theta(t)$ and the real noise $\epsilon(t)$ injected at timestep $t$. The second part is a Variational Lower Bound (VLB).
The Kullback-Leibler divergence $\mathcal{D}_{KL}$ is computed between two Gaussian distributions with $\tilde{\bm{\mu}}(x_t,x_0)=\frac{\sqrt{\alpha_t}(1-\bar{\alpha}_(t-1))\mathbf{x}_t)+\sqrt{\bar{\alpha}_{t-1}}(1-\alpha_t)\mathbf{x}_0}{1-\bar{\alpha_t}}$ $\tilde{\bm{\Sigma}}(t)=\frac{(1-\alpha_t)(1-\bar{\alpha}_{t-1})}{1-\bar{\alpha}_t}$ and $\bm{\mu}_\theta(\mathbf{x}_t,t)$ $\bm{\Sigma}_\theta(\mathbf{x}_t,t)$, respectively. 
$\gamma$ is a constant that increases the weight of small $t$ values when $\gamma$ is high. In our case, we found that $\gamma=2$ gives the best generation performance.

The training of the network is simply performed by randomly noising the samples at different timesteps $t$ using Equation \ref{eq:noising}. The sampling process, instead, start from a random Gaussian noise $\epsilon_T$ and after $T$ reverse steps generates the synthetic sample considering $q(\mathbf{x}_{t-1}|\mathbf{x}_t) \approx \mathcal{N}(\mathbf{x}_{t-1};\bm{\mu}_\theta(\mathbf{x}_t,t),\bm{\Sigma}_\theta(\mathbf{x}_t,t))$.
\section{EXPERIMENTAL SETUP AND RESULTS}
\label{sec:results}
In the context of hyperspectral DA, there are currently no well established metrics for assessing the quality of synthetically generated samples. So, we evaluate the effectiveness of the generated data by merging them with real data, performing a classification task and considering the classification accuracy as quality metric. Specifically, we compare the F1-score obtained by using only real data to the F1-score achieved with the inclusion of the augmented data. The difference between these scores serves as an indicator of the quality of synthetic samples generated by the DA methods.

\begin{figure}
\centering
\includegraphics[width=0.4\textwidth]{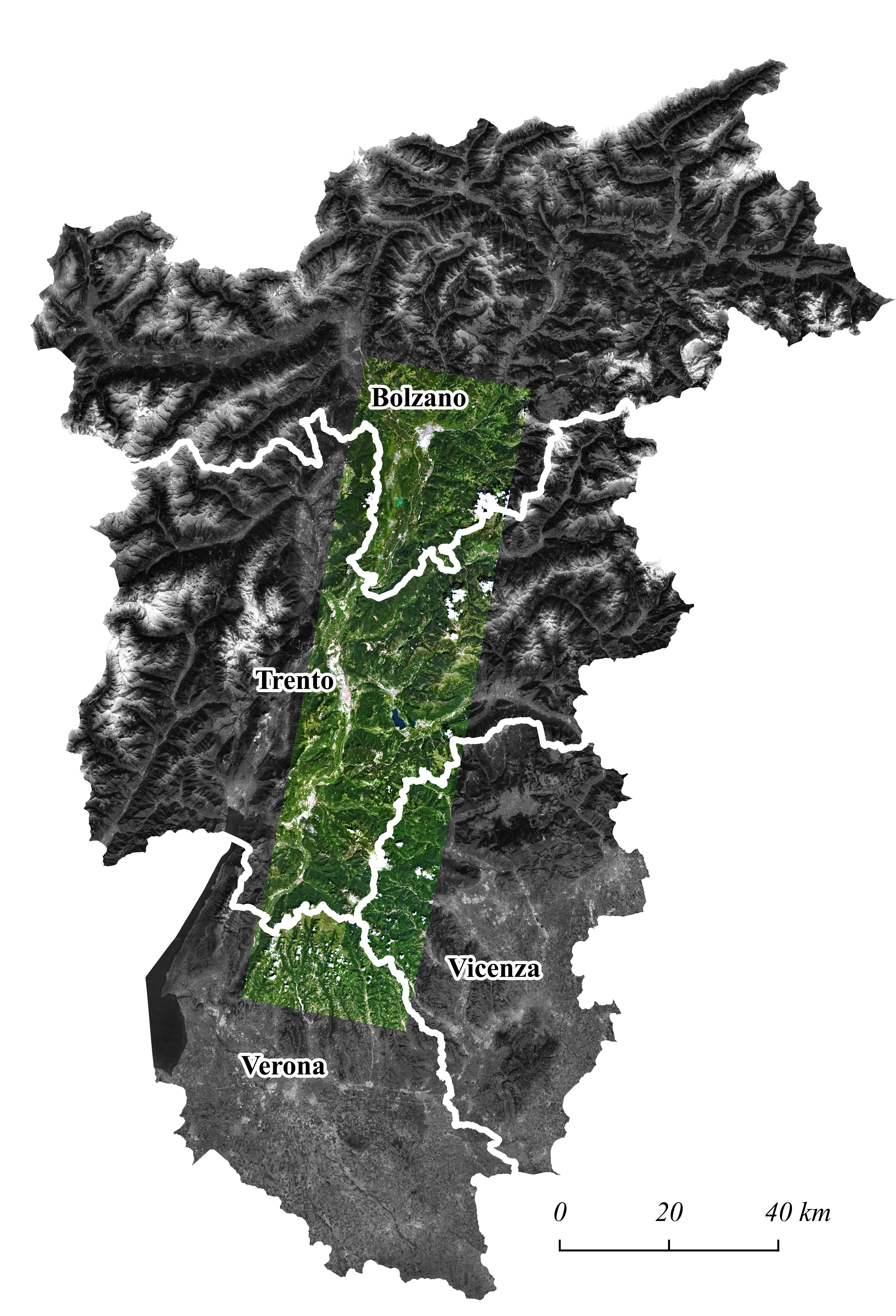}
\caption{Illustration of the study area where the considered set of PRISMA images is represented in true color }
\label{fig:aoi}
\end{figure}

The classification task considered involves categorizing forest types using a set of four PRISMA images acquired in the north-west region of Italy. The study area is depicted in Figure \ref{fig:aoi}. These images were acquired consecutively during the summer of 2023. The ground truth samples were obtained through both photointerpretation and ground surveys. The dataset consists of 8341 labeled samples, which are divided into the 10 forest categories reported in Table \ref{tab:dataset}.

The division of samples into training, validation, and test sets is carried out using a spatial clustering algorithm. Specifically, for each class, the samples are subdivided into four spatial clusters. Within each cluster, the samples are further spatially divided into three groups, corresponding to 60\%, 20\%, and 20\% of the samples relying on each cluster. These groups are then aggregated across all clusters to form the training, validation, and test sets, respectively. To test the methods on a scenario with very limited labeled data, the number of training samples is further reduced by randomly selecting 10\% of the group representing 60\% of the dataset. This splitting strategy aims to minimize the correlation between samples, thereby enhancing the capacity of evaluating the generalization capability of the model trained with the synthetic samples. Table \ref{tab:dataset} provides further details about the number of samples for each class. 

The experimental strategy involves two key steps. First, generative models are trained using the Tree-structured Parzen Estimator (TPE) optimization algorithm \cite{bergstra2011algorithms} over 200 trials with the goal of minimizing the loss function. The trained models are then used to generate synthetic samples, which are subsequently employed, combining them to real samples, to train and validate the network  in terms of classification accuracy. In the second step, the TPE algorithm is adopted to tune the learning rate and the weight decay of a 1-D CNN with five layers and skip connections that is used for classification task. The tuning process involves 100 trials of CNN training, with each trial using different values of learning rate and weight decay as selected by the optimization algorithm, in order to maximize classification accuracy on the validation set. The weights of the best-performing architecture are then used to classify the test set. To ensure the robustness and reliability of the results, the inference process is repeated with five different random seeds, and the average results are reported. During the tuning process on the validation set, we also optimize the parameters of both the transformation and pattern mixing methods.

The proposed DA approach is compared with several established methods to evaluate its effectiveness. The evaluated methods include: Jittering, which adds random Gaussian noise to spectral signatures; Scaling, which adjusts spectral signatures by a uniform scaling factor; and Magnitude Warping \cite{um2017data}, which involves selecting anchor points, applying cubic interpolation, and scaling these values to the spectral signature. Additionally, the comparison includes SMOTE \cite{chawla2002smote} and a GAN specifically designed for hyperspectral data.\cite{audebert2018generative} During the hyperparameter tuning of these methods we optimize: noise power for jittering and scaling, noise power and number of anchor points for magnitude warping, learning rate and weight decay for GAN and guided DM.

The classification results are presented in terms of the F1-Score percentage, as shown in Table \ref{tab:result}. Although the overall accuracy across methods is generally not very high due to the challenging classification task and the very limited number of real samples, the proposed method achieves the highest F1-Score, both in terms of average and weighted average across all classes. Most DA methods improve classification accuracy compared to using only real data, except for GAN and SMOTE. The lower performance of GAN may be attributed to difficulties in finding the convergence during the training, which can result in suboptimal generated data. The proposed method consistently achieved low training loss values across several trials, indicating a more stable and efficient training process compared to the GAN-based approach.

One-sample transformation methods also produced high-quality augmented samples, because the spectral variability in the spectral signature of a specific class is generally dominated by a scaling component \cite{borsoi2021spectral}. However, the proposed method excelled in generating synthetic samples under this condition and when more complex transformations dominated the spectral variability. Class-wise analysis shows that if we consider classes such as “Norway spruce” or “Black pines”, where sample differences are mainly scaling-related, and “Hornbeam and Hophornbeam”, where sample variations are more complex, the proposed method performed comparably to the most effective methods for each specific case demonstrating flexibility in generating synthetic samples. Finally, if we consider the “Other deciduous broadleaves” class, the use of only real data yielded better results due to the diverse spectral signatures associated to different trees species within this class.

\begin{table}
\centering
\caption{Classes and number of Samples of the PRISMA Forest Dataset}
\begin{tblr}{
  width = \linewidth,
  colspec = {Q[406]Q[171]Q[181]Q[177]},
  row{1} = {c},
  cell{2}{2} = {c},
  cell{2}{3} = {c},
  cell{2}{4} = {c},
  cell{3}{2} = {c},
  cell{3}{3} = {c},
  cell{3}{4} = {c},
  cell{4}{2} = {c},
  cell{4}{3} = {c},
  cell{4}{4} = {c},
  cell{5}{2} = {c},
  cell{5}{3} = {c},
  cell{5}{4} = {c},
  cell{6}{2} = {c},
  cell{6}{3} = {c},
  cell{6}{4} = {c},
  cell{7}{2} = {c},
  cell{7}{3} = {c},
  cell{7}{4} = {c},
  cell{8}{2} = {c},
  cell{8}{3} = {c},
  cell{8}{4} = {c},
  cell{9}{2} = {c},
  cell{9}{3} = {c},
  cell{9}{4} = {c},
  cell{10}{2} = {c},
  cell{10}{3} = {c},
  cell{10}{4} = {c},
  cell{11}{2} = {c},
  cell{11}{3} = {c},
  cell{11}{4} = {c},
  cell{12}{2} = {c},
  cell{12}{3} = {c},
  cell{12}{4} = {c},
  vlines,
  hline{1-2,12-13} = {-}{},
}
Classes                      & Tr. Samples & Val. Samples & Te. Samples                        \\
Larch and Swiss pine   & 61          & 219         & 219                                \\
Norway spruce                & 64          & 224         & 227                                \\
Fir                          & 56          & 219         & 216                                \\
Scots and Mountain pine & 68          & 214         & 215                                \\
Black pines                  & 24          & 83          & 81                                 \\
Beech                        & 37          & 135         & 136                                \\
Temperate oaks               & 35          & 102         & 102                                \\
Chestnut                     & 36          & 99          & 99                                 \\
Hornbeam and Hophornbeam     & 57          & 154         & 155                                \\
Other deciduous broadleaves  & 31          & 124         & 95                                 \\
Total                        & 469         & 1664        & 1670\textcolor[rgb]{0.8,0.8,0.8}{} 
\end{tblr}
\label{tab:dataset}
\end{table}

\hspace{1em}

\begin{table}
\centering
\caption{Classification Accuracy on Test set in terms of F1-Score \%}
\begin{tblr}{
  width = \linewidth,
  colspec = {Q[268]Q[89]Q[99]Q[91]Q[83]Q[80]Q[125]Q[85]},
  row{1} = {c},
  cell{2}{2} = {c},
  cell{2}{3} = {c},
  cell{2}{4} = {c},
  cell{2}{5} = {c},
  cell{2}{6} = {c},
  cell{2}{7} = {c},
  cell{2}{8} = {c},
  cell{3}{2} = {c},
  cell{3}{3} = {c},
  cell{3}{4} = {c},
  cell{3}{5} = {c},
  cell{3}{6} = {c},
  cell{3}{7} = {c},
  cell{3}{8} = {c},
  cell{4}{2} = {c},
  cell{4}{3} = {c},
  cell{4}{4} = {c},
  cell{4}{5} = {c},
  cell{4}{6} = {c},
  cell{4}{7} = {c},
  cell{4}{8} = {c},
  cell{5}{2} = {c},
  cell{5}{3} = {c},
  cell{5}{4} = {c},
  cell{5}{5} = {c},
  cell{5}{6} = {c},
  cell{5}{7} = {c},
  cell{5}{8} = {c},
  cell{6}{2} = {c},
  cell{6}{3} = {c},
  cell{6}{4} = {c},
  cell{6}{5} = {c},
  cell{6}{6} = {c},
  cell{6}{7} = {c},
  cell{6}{8} = {c},
  cell{7}{2} = {c},
  cell{7}{3} = {c},
  cell{7}{4} = {c},
  cell{7}{5} = {c},
  cell{7}{6} = {c},
  cell{7}{7} = {c},
  cell{7}{8} = {c},
  cell{8}{2} = {c},
  cell{8}{3} = {c},
  cell{8}{4} = {c},
  cell{8}{5} = {c},
  cell{8}{6} = {c},
  cell{8}{7} = {c},
  cell{8}{8} = {c},
  cell{9}{2} = {c},
  cell{9}{3} = {c},
  cell{9}{4} = {c},
  cell{9}{5} = {c},
  cell{9}{6} = {c},
  cell{9}{7} = {c},
  cell{9}{8} = {c},
  cell{10}{2} = {c},
  cell{10}{3} = {c},
  cell{10}{4} = {c},
  cell{10}{5} = {c},
  cell{10}{6} = {c},
  cell{10}{7} = {c},
  cell{10}{8} = {c},
  cell{11}{2} = {c},
  cell{11}{3} = {c},
  cell{11}{4} = {c},
  cell{11}{5} = {c},
  cell{11}{6} = {c},
  cell{11}{7} = {c},
  cell{11}{8} = {c},
  cell{12}{2} = {c},
  cell{12}{3} = {c},
  cell{12}{4} = {c},
  cell{12}{5} = {c},
  cell{12}{6} = {c},
  cell{12}{7} = {c},
  cell{12}{8} = {c},
  cell{13}{2} = {c},
  cell{13}{3} = {c},
  cell{13}{4} = {c},
  cell{13}{5} = {c},
  cell{13}{6} = {c},
  cell{13}{7} = {c},
  cell{13}{8} = {c},
  vlines,
  hline{1-2,12,14} = {-}{},
}
Classes                      & No Augmentation       & Proposed       & GAN & Jittering & Scaling        & Magn. Warp.    & SMOTE          \\
Larch and Swiss pine   & 80.58          & \textbf{89.19} & 77.06  & 81.32     & 86.98          & 83.13          & 72.97          \\
Norway spruce                & 72.77          & 74.49          & 76.92  & 71.67     & \textbf{78.03} & 72.10          & 65.77          \\
Fir                          & 76.55          & 76.54          & 76.81  & 72.81     & 76.19          & \textbf{79.73} & 71.20          \\
Scots and Mountain pine & 87.27          & \textbf{91.86} & 86.36  & 91.80     & 90.86          & 87.24          & 86.36          \\
Black pines                  & 85.53          & 90.67          & 78.79  & 90.90     & \textbf{93.00} & 91.34          & 85.71          \\
Beech                        & 80.97          & \textbf{83.33} & 78.23  & 81.23     & 74.55          & 76.38          & 78.08          \\
Temperate oaks               & 36.46          & \textbf{46.49} & 33.33  & 46.15     & 44.04          & 44.65          & 40.38          \\
Chestnut                     & 61.86          & 68.82          & 65.56  & 63.92     & 71.59          & \textbf{74.71} & 66.67          \\
Hornb. and Hophornb.     & 88.82          & 89.18          & 75.70  & 92.74     & 72.14          & 76.19          & \textbf{95.03} \\
Other dec. broadleaves  & \textbf{48.76} & 40.89          & 31.07  & 48.67     & 45.95          & 47.26          & 45.33          \\
Average                      & 71.96          & \textbf{75.15} & 67.98  & 74.12     & 73.33          & 73.28          & 70.75          \\
Weighted Average             & 74.35          & \textbf{77.39} & 71.14  & 75.84     & 75.51          & 74.94          & 72.08          
\end{tblr}
\label{tab:result}
\end{table}
\section{CONCLUSION}
\label{sec:CONCLUSION}
In this paper, we   have introduced a DA technique based on a DM to generate synthetic hyperspectral samples. The core of this method is a lightweight transformer network, with specific adjustments to the variance scheduler and loss function,  which enable fast and effective convergence during training, even with a limited number of labeled samples. We evaluated the effectiveness of the proposed method by comparing it with established DA techniques in a forest classification task involving multiple forest types and  hyperspectral PRISMA images. Our method consistently achieved superior or comparable performance across all classes, resulting in higher average accuracy and weighted accuracy. Additionally, the proposed approach demonstrated a fast and stable training process without any evidence of mode collapse behavior, making it particularly suitable for DA tasks, where achieving such characteristics is generally challenging with GAN-based methods. This study provides evidence of the effectiveness of diffusion-based DA for hyperspectral data. Future work will focus on further refining the method to better adapt it to the characteristics of hyperspectral data to enhance its overall performance.

\acknowledgments 
This activity has been supported by Italian Space Agency in the AFORISMA project (PRISMA SCIENZA call DC-UOT-2019-061).

\bibliography{report} 

\begin{thebibliography}{10}

\bibitem{fassnacht2016review}
Fassnacht, F.~E., Latifi, H., Stere{\'n}czak, K., Modzelewska, A., Lefsky, M., Waser, L.~T., Straub, C., and Ghosh, A., ``Review of studies on tree species classification from remotely sensed data,'' {\em Remote sensing of environment}~{\bf 186},  64--87 (2016).

\bibitem{guarini2017overview}
Guarini, R., Loizzo, R., Longo, F., Mari, S., Scopa, T., and Varacalli, G., ``Overview of the prisma space and ground segment and its hyperspectral products,'' in [{\em 2017 IEEE International Geoscience and Remote Sensing Symposium (IGARSS)}{\nolinebreak\hspace{0.1em}]},   431--434, IEEE (2017).

\bibitem{rodarmel2002principal}
Rodarmel, C. and Shan, J., ``Principal component analysis for hyperspectral image classification,'' {\em Surveying and Land Information Science}~{\bf 62}(2),  115--122 (2002).

\bibitem{zabalza2016novel}
Zabalza, J., Ren, J., Zheng, J., Zhao, H., Qing, C., Yang, Z., Du, P., and Marshall, S., ``Novel segmented stacked autoencoder for effective dimensionality reduction and feature extraction in hyperspectral imaging,'' {\em Neurocomputing}~{\bf 185},  1--10 (2016).

\bibitem{kingma2013auto}
Kingma, D.~P. and Welling, M., ``Auto-encoding variational bayes,'' {\em arXiv preprint arXiv:1312.6114}  (2013).

\bibitem{goodfellow2014generative}
Goodfellow, I., Pouget-Abadie, J., Mirza, M., Xu, B., Warde-Farley, D., Ozair, S., Courville, A., and Bengio, Y., ``Generative adversarial nets,'' {\em Advances in neural information processing systems}~{\bf 27} (2014).

\bibitem{ho2020denoising}
Ho, J., Jain, A., and Abbeel, P., ``Denoising diffusion probabilistic models,'' {\em Advances in neural information processing systems}~{\bf 33},  6840--6851 (2020).

\bibitem{dhariwal2021diffusion}
Dhariwal, P. and Nichol, A., ``Diffusion models beat gans on image synthesis,'' {\em Advances in neural information processing systems}~{\bf 34},  8780--8794 (2021).

\bibitem{liu2024diffusion}
Liu, Y., Yue, J., Xia, S., Ghamisi, P., Xie, W., and Fang, L., ``Diffusion models meet remote sensing: Principles, methods, and perspectives,'' {\em arXiv preprint arXiv:2404.08926}  (2024).

\bibitem{li2018data}
Li, W., Chen, C., Zhang, M., Li, H., and Du, Q., ``Data augmentation for hyperspectral image classification with deep cnn,'' {\em IEEE Geoscience and Remote Sensing Letters}~{\bf 16}(4),  593--597 (2018).

\bibitem{chawla2002smote}
Chawla, N.~V., Bowyer, K.~W., Hall, L.~O., and Kegelmeyer, W.~P., ``Smote: synthetic minority over-sampling technique,'' {\em Journal of artificial intelligence research}~{\bf 16},  321--357 (2002).

\bibitem{wang2019hyperspectral}
Wang, C., Zhang, L., Wei, W., and Zhang, Y., ``Hyperspectral image classification with data augmentation and classifier fusion,'' {\em IEEE Geoscience and Remote Sensing Letters}~{\bf 17}(8),  1420--1424 (2019).

\bibitem{wang2021data}
Wang, W., Liu, X., and Mou, X., ``Data augmentation and spectral structure features for limited samples hyperspectral classification,'' {\em Remote Sensing}~{\bf 13}(4),  547 (2021).

\bibitem{haut2019hyperspectral}
Haut, J.~M., Paoletti, M.~E., Plaza, J., Plaza, A., and Li, J., ``Hyperspectral image classification using random occlusion data augmentation,'' {\em IEEE Geoscience and Remote Sensing Letters}~{\bf 16}(11),  1751--1755 (2019).

\bibitem{zhu2018generative}
Zhu, L., Chen, Y., Ghamisi, P., and Benediktsson, J.~A., ``Generative adversarial networks for hyperspectral image classification,'' {\em IEEE Transactions on Geoscience and Remote Sensing}~{\bf 56}(9),  5046--5063 (2018).

\bibitem{he2022hypervitgan}
He, Z., Xia, K., Ghamisi, P., Hu, Y., Fan, S., and Zu, B., ``Hypervitgan: Semisupervised generative adversarial network with transformer for hyperspectral image classification,'' {\em IEEE Journal of Selected Topics in Applied Earth Observations and Remote Sensing}~{\bf 15},  6053--6068 (2022).

\bibitem{audebert2018generative}
Audebert, N., Le~Saux, B., and Lef{\`e}vre, S., ``Generative adversarial networks for realistic synthesis of hyperspectral samples,'' in [{\em IGARSS 2018-2018 IEEE International Geoscience and Remote Sensing Symposium}{\nolinebreak\hspace{0.1em}]},   4359--4362, IEEE (2018).

\bibitem{hennessy2021generative}
Hennessy, A., Clarke, K., and Lewis, M., ``Generative adversarial network synthesis of hyperspectral vegetation data,'' {\em Remote Sensing}~{\bf 13}(12),  2243 (2021).

\bibitem{zhang2023features}
Zhang, M., Wang, Z., Wang, X., Gong, M., Wu, Y., and Li, H., ``Features kept generative adversarial network data augmentation strategy for hyperspectral image classification,'' {\em Pattern Recognition}~{\bf 142},  109701 (2023).

\bibitem{wang2019caps}
Wang, X., Tan, K., Du, Q., Chen, Y., and Du, P., ``Caps-triplegan: Gan-assisted capsnet for hyperspectral image classification,'' {\em IEEE Transactions on Geoscience and Remote Sensing}~{\bf 57}(9),  7232--7245 (2019).

\bibitem{croitoru2023diffusion}
Croitoru, F.-A., Hondru, V., Ionescu, R.~T., and Shah, M., ``Diffusion models in vision: A survey,'' {\em IEEE Transactions on Pattern Analysis and Machine Intelligence}  (2023).

\bibitem{rombach2022high}
Rombach, R., Blattmann, A., Lorenz, D., Esser, P., and Ommer, B., ``High-resolution image synthesis with latent diffusion models,'' in [{\em Proceedings of the IEEE/CVF conference on computer vision and pattern recognition}{\nolinebreak\hspace{0.1em}]},   10684--10695 (2022).

\bibitem{chen2023spectraldiff}
Chen, N., Yue, J., Fang, L., and Xia, S., ``Spectraldiff: A generative framework for hyperspectral image classification with diffusion models,'' {\em IEEE Transactions on Geoscience and Remote Sensing}  (2023).

\bibitem{sigger2024unveiling}
Sigger, N., Vien, Q.-T., Nguyen, S.~V., Tozzi, G., and Nguyen, T.~T., ``Unveiling the potential of diffusion model-based framework with transformer for hyperspectral image classification,'' {\em Scientific Reports}~{\bf 14}(1),  8438 (2024).

\bibitem{zhou2024exploring}
Zhou, J., Sheng, J., Ye, P., Fan, J., He, T., Wang, B., and Chen, T., ``Exploring multi-timestep multi-stage diffusion features for hyperspectral image classification,'' {\em IEEE Transactions on Geoscience and Remote Sensing}  (2024).

\bibitem{sousa2024data}
Sousa, T., Ries, B., and Guelfi, N., ``Data augmentation in earth observation: A diffusion model approach,'' {\em arXiv preprint arXiv:2406.06218}  (2024).

\bibitem{wu2023high}
Wu, Z., Qian, J., Zhang, M., Cao, Y., Wang, T., and Yang, L., ``High-confidence sample augmentation based on label-guided denoising diffusion probabilistic model for active deception jamming recognition,'' {\em IEEE Geoscience and Remote Sensing Letters}  (2023).

\bibitem{nichol2021improved}
Nichol, A.~Q. and Dhariwal, P., ``Improved denoising diffusion probabilistic models,'' in [{\em International conference on machine learning}{\nolinebreak\hspace{0.1em}]},   8162--8171, PMLR (2021).

\bibitem{peebles2023scalable}
Peebles, W. and Xie, S., ``Scalable diffusion models with transformers,'' in [{\em Proceedings of the IEEE/CVF International Conference on Computer Vision}{\nolinebreak\hspace{0.1em}]},   4195--4205 (2023).

\bibitem{bergstra2011algorithms}
Bergstra, J., Bardenet, R., Bengio, Y., and K{\'e}gl, B., ``Algorithms for hyper-parameter optimization,'' {\em Advances in neural information processing systems}~{\bf 24} (2011).

\bibitem{um2017data}
Um, T.~T., Pfister, F.~M., Pichler, D., Endo, S., Lang, M., Hirche, S., Fietzek, U., and Kuli{\'c}, D., ``Data augmentation of wearable sensor data for parkinson’s disease monitoring using convolutional neural networks,'' in [{\em Proceedings of the 19th ACM international conference on multimodal interaction}{\nolinebreak\hspace{0.1em}]},   216--220 (2017).

\bibitem{borsoi2021spectral}
Borsoi, R.~A., Imbiriba, T., Bermudez, J. C.~M., Richard, C., Chanussot, J., Drumetz, L., Tourneret, J.-Y., Zare, A., and Jutten, C., ``Spectral variability in hyperspectral data unmixing: A comprehensive review,'' {\em IEEE geoscience and remote sensing magazine}~{\bf 9}(4),  223--270 (2021).

\end{thebibliography}
\bibliographystyle{spiebib} 

\end{document}